\documentclass[a4paper]{article}

\usepackage{INTERSPEECH2020}

\usepackage[table,xcdraw]{xcolor}
\usepackage{multicol}
\usepackage{multirow}
\usepackage{amsmath}
\usepackage[font=small,skip=5pt]{caption}
\usepackage{tabularx, makecell}
\usepackage{stfloats}

\usepackage{graphicx}
\usepackage{subcaption}
\title{The ``Sound of Silence" in EEG - Cognitive voice activity detection}

\name{Rini A Sharon$^1$, Hema A Murthy$^1$}

\address{
  $^1$Indian Institute of Technology, Madras} 
\email{ee15d210@smail.iitm.ac.in, hema@cse.iitm.ac.in}

\begin{document}

\maketitle
\begin{abstract}
Speech cognition bears potential application as a brain computer interface that can improve the quality of life for the otherwise communication impaired people. While speech and resting state EEG are popularly studied, here we attempt to explore a ``non-speech"(NS) state of brain activity corresponding to the silence regions of speech audio. Firstly, speech perception is studied to inspect the existence of such a state, followed by its identification in speech imagination. Analogous to how voice activity detection is employed to enhance the performance of speech recognition, the EEG state activity detection protocol implemented here is applied to boost the confidence of imagined speech EEG decoding. Classification of speech and NS state is done using two datasets collected from laboratory-based and commercial-based devices. The state sequential information thus obtained is further utilized to reduce the search space of imagined EEG unit recognition. Temporal signal structures and topographic maps of NS states are visualized across subjects and sessions. The recognition performance and the visual distinction observed demonstrates the existence of silence signatures in EEG.

\end{abstract}
\noindent\textbf{Index Terms}: Speech-EEG silence recognition, Brain-computer interface, Two level dynamic programming

\section{Introduction}

Analysis and decoding of brain signals while the subject imagines speech has become an important topic of interest\cite{nguyen2017inferring, hashim2017word}, because of its applicability as a Brain Computer Interface(BCI) for the speech and motor impaired individuals\cite{lazarou2018eeg, brumberg2018brain}.
Widely deployable user-friendly BCI devices work on non-invasive methods\cite{waldert2016invasive} of brain data collection such as   Electroencephalogram(EEG) that captures the electrical activity of the brain using electrodes placed on the scalp of the subjects\cite{subha2010eeg}.


While speech-EEG based BCI systems commonly focus on speech-unit classification(where vowels, syllables, words and phrases are considered as units\cite{min2016vowel, hashim2017word, dash2020decoding, 9056076}), the speech-silence portions associated with these units lack analysis. Since speech and EEG are both well correlated, temporally informative signals, studying the inherent existence of silence signatures in the brain is both meaningful and interesting\cite{horton2014envelope, vanthornhout2019effect}. The broad objective of this paper therefore lies in determining if speech-silence signatures exist in brain signals, if so, do they exhibit unique temporal patterns, in which case, can these distinct signatures be used to improve the decoding of speech-EEG. 



The rest of the paper includes Section \ref{sec:motiv} which discusses the motivation to pursue this line of thought and the previous works that are relatable to our objective. This is followed by Section \ref{sec:data} which provide an overview of the data collection process and the databases involved. Section \ref{sec:prof} then explains the proposed framework and its associated experimental details.  The results are then discussed in Section \ref{sec:result} and the work is summarized in Section \ref{sec:conc}. 

\begin{figure}[t]
  \centering
  \includegraphics[width=\linewidth,trim={0cm 0cm 0cm 0cm},clip]{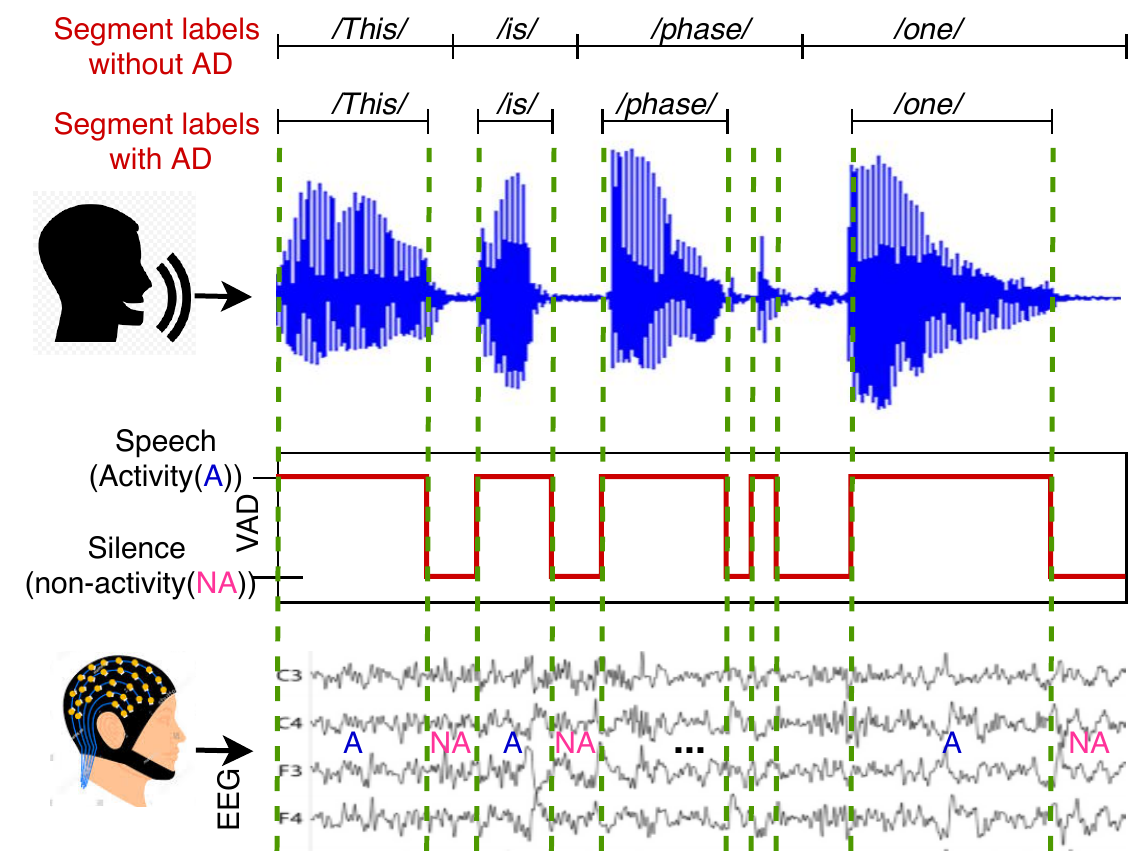}
  \caption{Segmentation using Voice Activity Detection}
  \label{fig:motiv}
  \vspace{-0.4cm}
\end{figure}

\section{Motivation and Related Work}\label{sec:motiv}


Speech signals consist of three broadly classified regions, namely, voiced regions where the vocal cords vibrate to produce sounds, unvoiced regions produced by sounds of whisper or aspiration, and silence regions where no speech is produced\cite{qi1993voiced}. The energy and amplitude of the signal in the silence region is therefore very low. Since no data is being captured during silence, identifying and removing silence regions significantly reduces the processing time of the system and improves the performance of the speech recognizer\cite{asadullah2017silence}. For such purposes, voice activity detection(VAD) algorithms are employed\cite{ding2019personal, drugman2015voice}. VAD deactivates the system operation during silence regions thus avoiding the unwanted processing and storage of non-informative frames. 

Works dealing with imagined speech EEG decoding so far assume the whole segment of the EEG data as belonging to a particular unit class. However, considering the nature of speech signals and its reflection on cognitive processing, there could exist non-activity states before, between and after the speech units.  Consider VAD for the case of phrase-level speech unit identification as depicted in Figure \ref{fig:motiv}. The non-activity states(silence) in speech do not directly correspond to the target class itself but have a generic non-speech structure common across all the data. Therefore, drawing a parallelism for EEG, modelling the system by mapping the whole segment to a single target class would not be as effective as sub dividing the segment into activity and non-activity portions and modelling them independently.

As far as our knowledge goes, this is the first time a silence or non-speech mapping to EEG signals is considered for analysis. Throughout this work, we define a ``non-speech"(NS) state in EEG as a state that does not correspond to brain cognitive speech activity. A noteworthy point of comparison here is the difference between rest state EEG as studied in many works and the NS state studied here\cite{scally2018resting, bai2017review}. Resting state EEG is recorded while the subject is not performing any voluntary activity whereas the NS state EEG corresponds to temporal regions of the EEG recorded while the subject is consciously performing an activity. Unlike silence in speech, the NS state may not have consistent characteristics in different parts of its occurrences. Contrary to how silence segments are visually distinguishable and easily identifyable in speech signals, the existence of a mapping between speech silences and their corresponding brain signals is still unclear. This therefore is the primary motivation for the study of NS states in EEG. 

\begin{figure}[t]
  \centering
  \includegraphics[width=\linewidth,trim={0cm 0cm 0cm 0cm},clip]{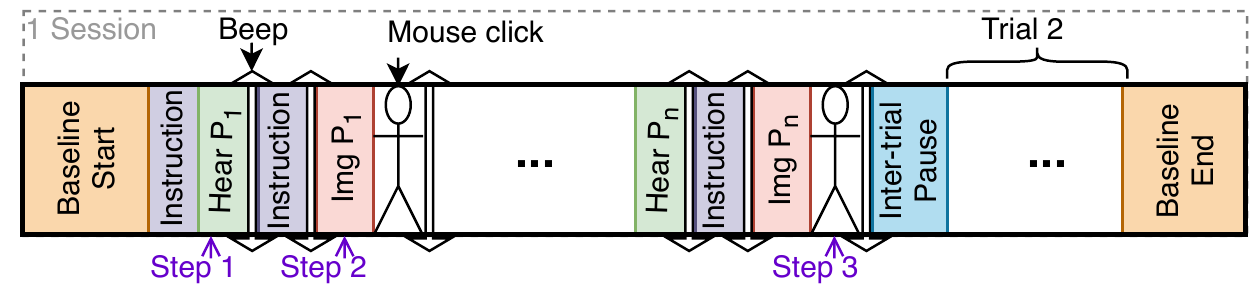}
  \vspace{-1cm}
  \label{fig:timeline}
\end{figure}

\begin{figure*}[bp]
  \centering
  \includegraphics[width=\linewidth,trim={0cm 0cm 0cm 0cm},clip]{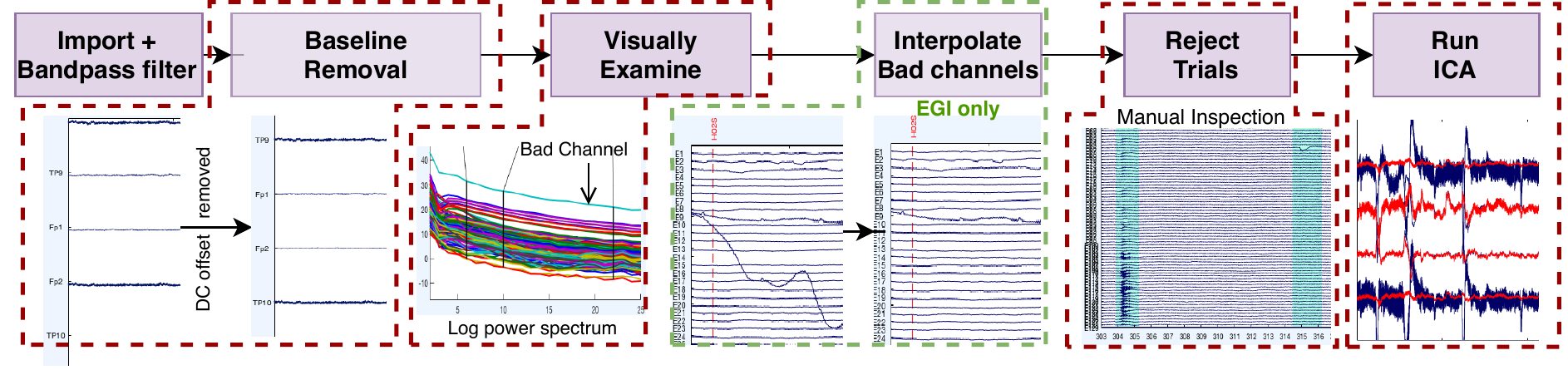}
  \caption{Pre-processing}
  \label{fig:pp}
  \vspace{-0.4cm}
\end{figure*}

Much of the previous EEG based imagined speech recognition studies require high channel density EEG devices with the added inconvenience of conductive gel based electrode placement. Despite the lack of quality and resolution of the signal,  the affordability and portability of consumer EEG headsets like MUSE\cite{muse} make them attractive alternatives. Research using the MUSE device so far mainly focuses on tasks like meditation \cite{surangsrirat2015analysis}, quantifying human attention abilities \cite{przegalinska2018muse}, detecting user engagement\cite{abujelala2016brain} and mental vigilance monitoring \cite{armanfard2016vigilance}. Other BCI related MUSE work include ERP analysis\cite{krigolson2017choosing, krigolson2017using} and cognitive state detection\cite{bashivan2016mental, bird2019mental}. Here we introduce an attempt to improve decoding of imagined speech EEG using VAD by employing the MUSE device alongside a 128 channel EGI device.

\section{EEG Data Acquisition} \label{sec:data}
The Ethics Committee of the Indian Institute of Technology Madras approved this study. All volunteers were healthy subjects in the age group 21-30 years. The subjects were briefed about the experimental protocol and an initial demo of the structure of the experiment was given. Post this a written consent was obtained, the EGI cap or the Muse headband was mounted and the subject was comfortably seated in an an-echoic chamber. They were instructed to keep their eyes closed and consciously restrict other movements to get minimal artifact intervention. The electrode impedance were monitored and kept below an acceptable threshold. Speakers placed 4 feet in front of the subject were used to play the audio instructions and input cues. The input audio sentences were recorded by 1 male and 1 female volunteer. 16 subjects volunteered to provide EEG data for Dataset-1, of which 4 appeared for 2 sessions each. Dataset-2 involved 8 subjects with 3 sessions each.

\subsection{Dataset 1}
This dataset is collected using a 128 channel EGI Geodesic net at a sampling rate of 250Hz\cite{egi}. 
The timeline of the experiment is as shown in Figure \ref{fig:timeline}. The subject is expected to perform four steps, a common baseline relaxation step and three phrase specific steps for each of the input phrases in a trial. A total of 19 daily use phrases were played as inputs in a single trial. For every phrase, step 1 requires the subject to pay attention by passively listening to the audio played. Step 2 involves the subject imagining as if he/she is speaking the sentence that was just played. Once done, the subject is asked to indicate the end of the imagination action by means of a mouse click in step 3. The 28 phrases(19 from two speakers) are chosen in a random order and played once in every trial. Two such trials are conducted in each session.  

\subsection{Dataset 2}

For this dataset, EEG signals were recorded using the InteraXon Muse EEG headset, a non-invasive, low cost, commercially available wearable device that does not require the application of a conductive gel or solution\cite{muse}. Muse provides four channel data from two electrodes placed on the forehead and two placed on the scalp behind the ears. It produces bipolar readings using the electrode
located analogous to Fpz(center of forehead) as the reference. The other 4 data channels in Muse correspond to Fp1, Fp2, TP9, and TP10 locations in the 10-20 system of electrode placement\cite{jatupaiboon2013real}. The signals are generated at a sampling rate of 1KHz. Proper connectivity of all the four channels was continuously monitored.  Muse EEG data is transferred to an Android mobile phone using Bluetooth 2.1 + EDR and the Muse-player tool was used to convert the data format into MATLAB readable forms. MATLAB was also used for stimulus presentation and for marking the timestamps. The system time and the mobile application clock time was recorded at the beginning, middle and the end of the experiment so that the latency due to connectivity and data transfer could be synced offline. The experimental timeline followed for Dataset 1 is adopted here, except that a subset of 5 phrases(containing 9 words in total) out of the 19 is chosen as inputs. 

\subsection{Pre-processing}
Following EEG data collection, the data was pre-procesed using MATLAB EEGLab toolkit\cite{delorme2004eeglab} in stages as described in Figure \ref{fig:pp}. First, a band pass Butterworth filter with a passband of 0.1 Hz to 60 Hz was applied in addition to a 50 Hz notch filter to eliminate AC interference. Then a baseline correction was done to remove the DC offset of the channels. The log power spectra of all the channels were visualized to identify bad channels and the channels so identified, if any, were interpolated(in our case this was done only for the EGI 128 channel data). Manual inspection of the data across time was done to eliminate trials containing subject induced artifacts. Finally Independent component analysis was performed and the top one and three components were removed for Muse and EGI respectively. The data is then segmented using the markers used to denote the onset and completion of the stimulus. These segments are considered as independent training and testing instances for modelling.

\section{Proposed Framework}\label{sec:prof}

\subsection{General Experimental set-up}

For a fair comparison between the Muse and EGI system performances, classification of trials belonging to the 5 phrases common to both datasets were taken for experimentation. Three test scenarios were formulated as follows:
\begin{enumerate}
    \item Intra-session: Trials from the same sessions are split into train and test sets in a disjoint 7:3 ratio.
    \item Inter-session: Sessions seen in train-set are excluded from test-set. Only subjects having recordings of multiple sessions were analyzed. 
    \item Inter-subject: Leave-one-subject out testing was performed. $N$-fold cross validation results are reported, were $N$ is the number of subjects in the database. 
\end{enumerate}

The EEG VAD module is designed and trained to predict whether a input frame belongs to the speech(S) state or a non-speech(NS) state. These NS states correspond to the EEG data during silence intervals in speech.
In a particular segment, three types of NS states are modelled. The NS$_b$ state denotes the silence portion at the beginning of the segment and the NS$_e$ state denotes the silence at the end of the segment. Since auditory scientists report that accurate perception of syllables, and the words they compose, highly depend on the silence gaps between them\cite{held2012perception, myers1981cognitive, dorman1979some}, the state NS$_i$ is defined, which denotes the  silence between the words/syllables/units in the segment. Both the passive audition EEG and the imagine EEG are analyzed. 

\subsection{Training and Testing Methodology}


The training stage begins with the segmentation of the EEG trials, pre-processing them and extracting the required features. Once the frame-level features are extracted, our proposed method suggests a hierarchical classification strategy. An activity detection(AD) classifier is first modelled followed by a speech unit detection classifier. A VAD module is employed for the input speech signal with the objective of determining whether an input frame belongs to speech activity or not. Depending on the output of the VAD algorithm implemented using \cite{cvx}, a frame-level label modification is done. AD models are now trained considering a 4-class set-up, with S, NS$_b$, NS$_i$ and NS$_e$ being the four classes. Alongside this, another set of models for the individual speech units are also trained using the segment boundaries hypothesised by the VAD module. Speech ``units" hereon-wards refer to the speech segments annotated by the VAD module and can be word-level or syllable-level demarcations(W/S units). 

\begin{figure}[t]
  \centering
  \includegraphics[width=\linewidth,trim={0cm 0cm 0cm 0cm},clip]{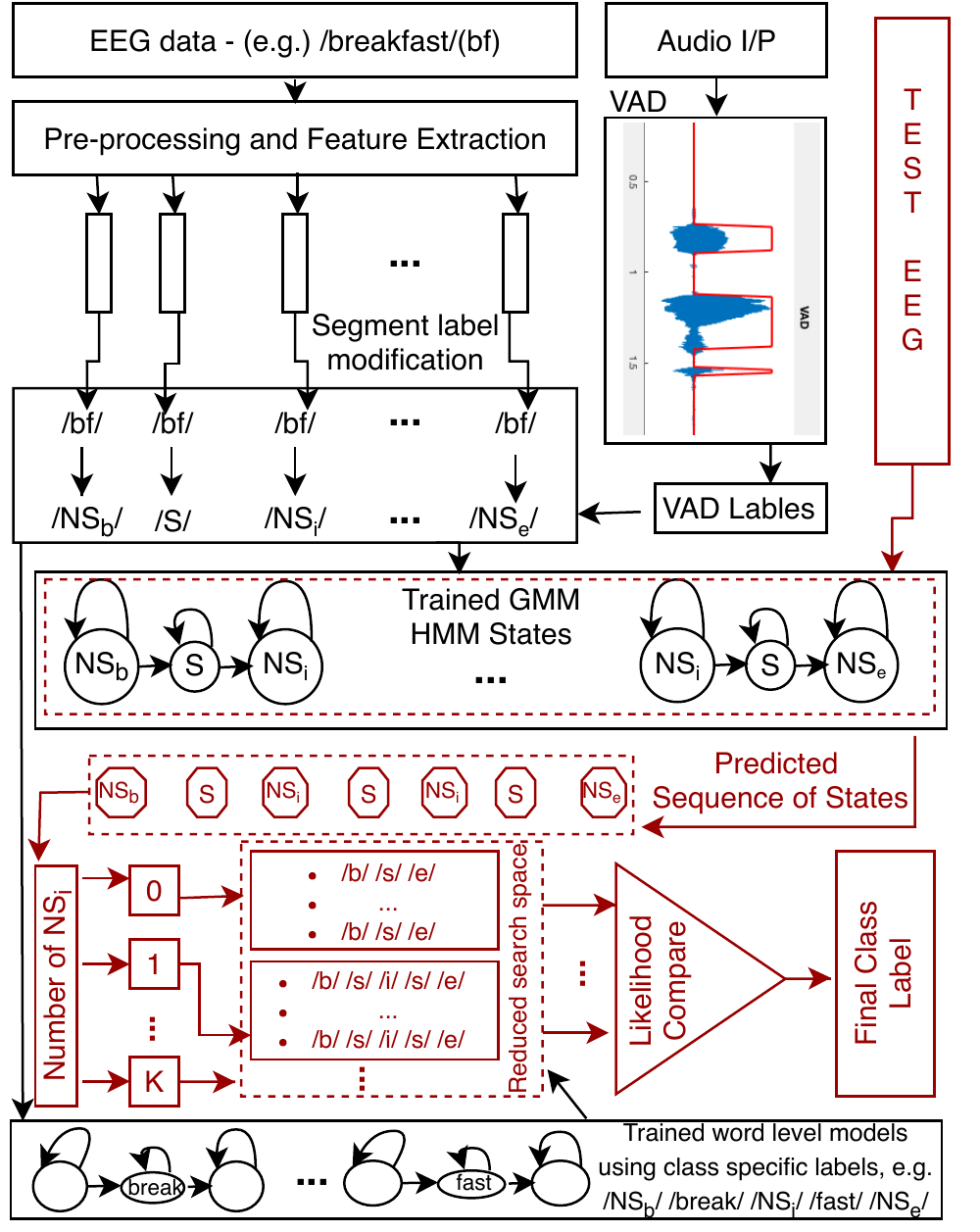}
  \caption{Proposed Architecture}
  \label{fig:arc}
  \vspace{-0.4cm}
\end{figure}

During testing, the test EEG segment is passed through the VAD classifier which outputs a sequence of states. Depending on the number of intermediate silence states, the search space is reduced and the likelihoods are only compared for a subset of classes having a matching number of NS$_i$ states. The final label is assigned based on the outputs of these subsets as depicted in Figure \ref{fig:arc}. 

\subsection{Feature Extraction and Classification Module}
Short term time domain processing is a common approach followed to extract meaningful signal characteristics in the field of temporally informative signals like speech and EEG\cite{nandhini2014voiced,sanei2013eeg}. In this work, we adopt the short term energy(STE) feature extraction technique\cite{sharon2019empirical}. The STE is calculated in MATLAB \cite{MATLAB:R2020a} using Equation \ref{eqn:ste}, where $U$ is the STE, $W$ is a hamming window function of length 125 samples and $E$ is the input EEG signal. 

\begin{equation} \label{eqn:ste}
U(n) = \sum_{m} [E(m)W(n-m)]^{2}
\end{equation}

A Gaussian Mixture based Hidden Markov model(GMM-HMM) is employed for classification using the Kaldi Toolkit\cite{kipyatkova2016dnn}. The GMMs model the means and variances of the data instances and the HMMs perform temporal modelling using transition and emission probabilities. Hyper-parameters such as number of mixtures(average of 3 per state) and number of HMM states were tuned. 40 update iterations were run and the lattice beam width for decoding was varied between 1 to 4 in the Viterbi algorithm. For the first level of classification, a boost silence probability(to detect the NS state) of 1.28 was applied. 

\subsection{Visualization - CED}
To visualize the NS segments and to inspect if they have distinct signatures, we use a compress-expand dynamic time warping(CED) method as described in \cite{abdulla2003cross}. The S/NS$_{b,i,e}$ classification boundaries of all the trials are obtained and the data corresponding to the boundaries are extracted. Since these segments are of differing lengths, CED is performed to make them equilength. The mean of these equilength segments and their variances are plotted across time to analyze the temporal structures unique to a particular unit class. 

\begin{table*}[t] 
\centering
\caption{Unit classification accuracies- highest accuracy is highlighted in each block} \label{tab:1}
\begin{tabular}{|c|cccc|cccc|cccc|}
\hline
\rowcolor[HTML]{EFEFEF} 
\textbf{}                         & \multicolumn{4}{c|}{\cellcolor[HTML]{EFEFEF}\textbf{Intra-session}}                                                         & \multicolumn{4}{c|}{\cellcolor[HTML]{EFEFEF}\textbf{Inter-session}}                                                            & \multicolumn{4}{c|}{\cellcolor[HTML]{EFEFEF}\textbf{Inter-subject}}                                                             \\ \hline
\rowcolor[HTML]{C0C0C0} 
{\color[HTML]{000000} Model-type} & {\color[HTML]{000000} BL} & {\color[HTML]{000000} DNS} & {\color[HTML]{000000} DNS3} & {\color[HTML]{000000} HC}            & {\color[HTML]{000000} BL} & {\color[HTML]{000000} DNS} & {\color[HTML]{000000} DNS3} & {\color[HTML]{000000} HC}            & {\color[HTML]{000000} BL} & {\color[HTML]{000000} DNS} & {\color[HTML]{000000} DNS3} & {\color[HTML]{000000} HC}            \\
Muse                              & 40.2                      & 43.4                       & 41.7                        & {\color[HTML]{00009B} \textbf{50.3}} & 38.1                      & 39.8                       & 41.8                        & {\color[HTML]{00009B} \textbf{49.2}} & 23.4                      & 26.2                       & 24.3                        & 28.2                                 \\
EGI                               & 43.8                      & 45.6                       & 45.8                        & 49.4                                 & 42.6                      & 43.1                       & 42.4                        & 47.8                                 & 27.3                      & 28.2                       & 30.8                        & {\color[HTML]{00009B} \textbf{32.6}} \\
EGI-reduced                       & 37.9                      & 38.2                       & 40.1                        & 42.3                                 & 37.3                      & 36.4                       & 40.8                        & 42.9                                 & 24.3                      & 24.3                       & 25.7                        & 28.8                                 \\ \hline
\end{tabular}
\vspace{-0.4cm}
\end{table*}

\section{Results and Discussion}\label{sec:result}
All the classification protocols consider ``1-UER" as the metric for accuracy, where the unit error rate(UER) is calculated by summing the total number of insertions, deletions and substitutions in the decode output and dividing it by the total number of units in the database. The sequence of ground truth(GT) labels in each of the models for the imagined EEG are assumed to be the same as the immediately preceding heard EEG GT labels.
 
\begin{figure}[t]
\begin{subfigure}{\textwidth}
  \includegraphics[width=0.45\textwidth,trim={1cm 0.9cm 1.8cm 5cm},clip]{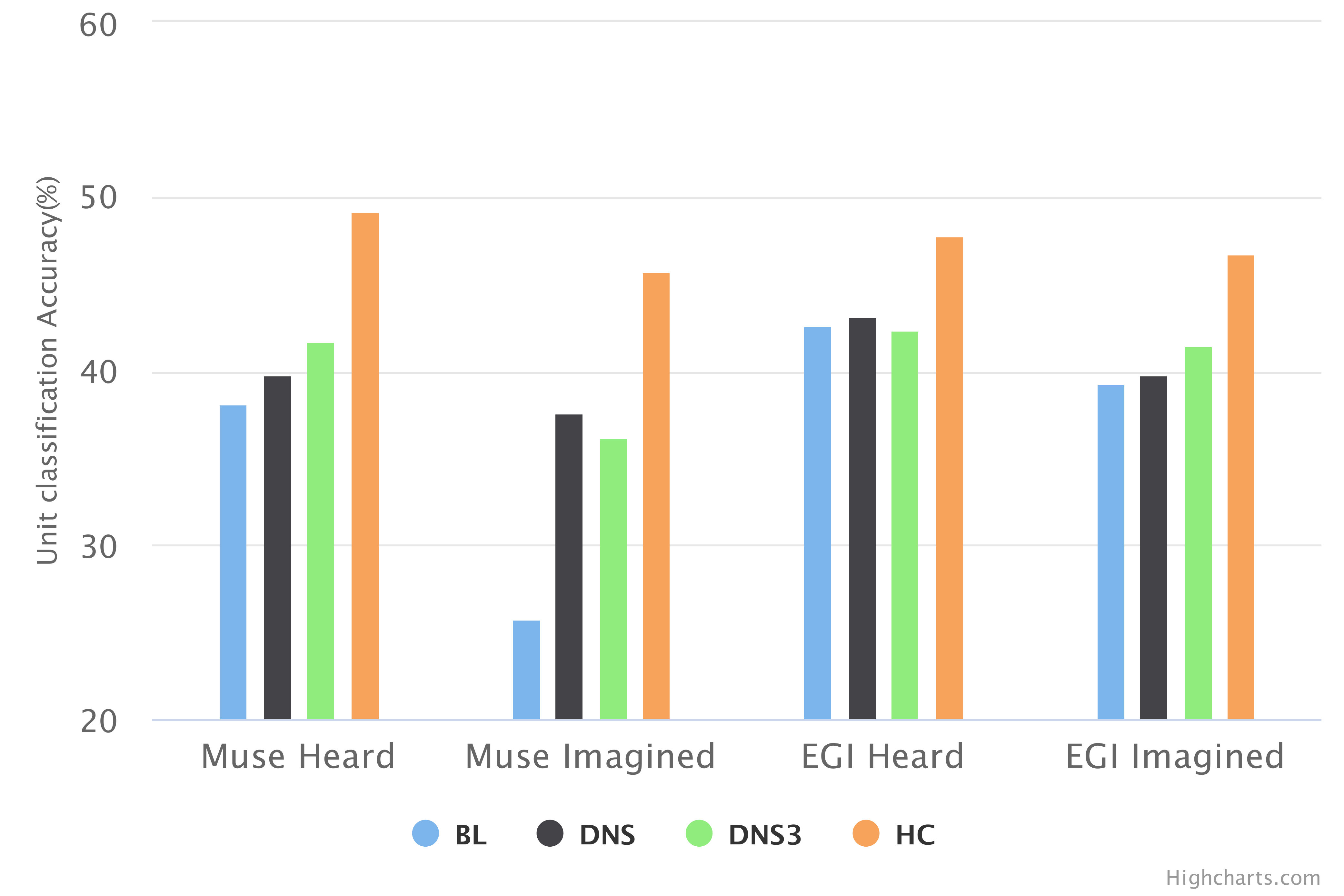}
  \end{subfigure}
\medskip
\begin{subfigure}{0.23\textwidth}
\centering
\includegraphics[width=\linewidth, trim={0cm 0cm 0.5cm 0cm},clip]{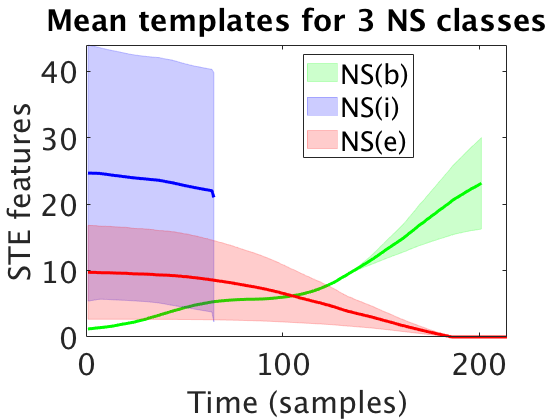}\quad
  \label{fig:11}
\end{subfigure}
\begin{subfigure}{0.23\textwidth}
\centering
\includegraphics[width=\linewidth, trim={0cm 0cm 0.5cm 0cm},clip]{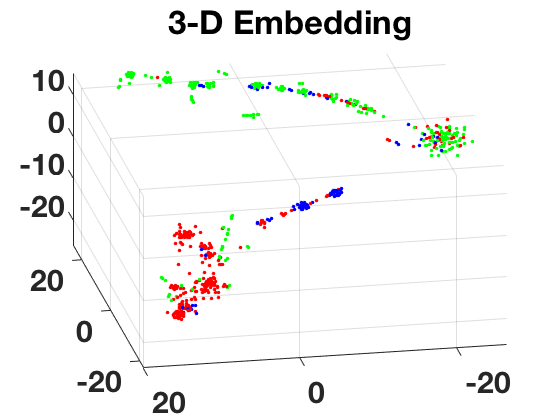}
\end{subfigure}
\caption{Unit classification accuracy and Feature visualization}
  \label{fig:vis}
  \vspace{-0.4cm}
\end{figure}

\subsection{Classification Results}

In the classification framework, our broad objective is to classify heard and imagined speech EEG W/S units. To achieve this we propose three model-level design schemes and compare it with the popularly followed baseline technique. The baseline(BL) is defined as a model trained and tested by giving the segment labels without AD as depicted in Figure \ref{fig:motiv} as the GT labels(dividing the entire segment into its constituent sequence of W/S-units alone). The three proposed models are:
\begin{enumerate}
    \item Direct classification of segments including the NS state to denote no-activity in the GT labels(DNS).
    \vspace{-0.1cm}
    \item Direct classification of segments including the 3 subdivisions of NS states(NS$_{b,i,e}$) in the GT labels(DNS3).
    \vspace{-0.1cm}
    \item Hierarchical classification(HC) using intermediate AD stage before final classification(proposed in Figure \ref{fig:arc}).
\end{enumerate}

Inclusion of NS state labels boost the recognition performance, as the data segments previously considered to model EEG speech units, now exclude the brain NS regions. The HC method performs best by giving an absolute average improvement of 7.8\% accuracy over the BL as shown in Figure \ref{fig:vis} for the inter-session case. The accuracy of the activity detection classifier is $\approx$76\%, suggesting that this module  efficiently captures discrimination. This better modelling could also be a result of larger amounts of effective data used to model these states(since every trial contributes at least one data instant to each class). 

For real-life applicability, we need to investigate whether our models capture generic EEG patterns across different session and users. Since our datasets include multiple subjects and sessions, we test for generalization using the intra/inter subject/session testing scenarios and report the results for heard EEG in Table \ref{tab:1}. Muse set-up functions better for the intra-subject cases and EGI scales well for the inter-subject experiments. To afford a better performance comparison between the EGI and the Muse Datasets, a reduced analysis on Dataset 2 was done(EGI-reduced). Only the 4 muse channels(TP9, TP10, Fp1, Fp2) were retained. These channels were then re-referenced considering Fpz as the reference electrode. As observed from Table \ref{tab:1}, EGI-reduced does not perform in same capacity as Muse, but follows the model-wise performance pattern.

\subsection{Topographic maps and Temporal structures}
\begin{figure}[t]
\begin{subfigure}{0.15\textwidth}
\centering
\includegraphics[width=\linewidth, trim={2cm 1.5cm 2cm 1cm},clip]{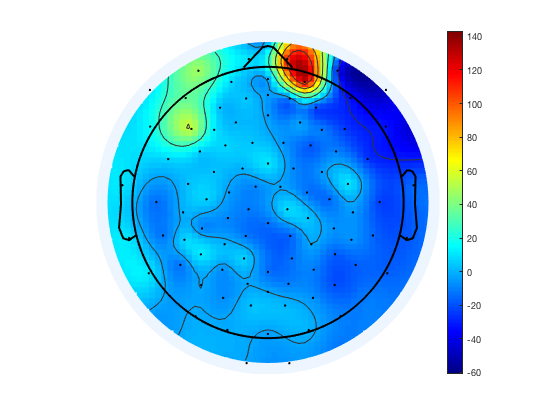}\quad
\caption{NS$_b$}
  \label{fig:11}
\end{subfigure}
\begin{subfigure}{0.15\textwidth}
\centering
\includegraphics[width=\linewidth, trim={2cm 1.5cm 2cm 1cm},clip]{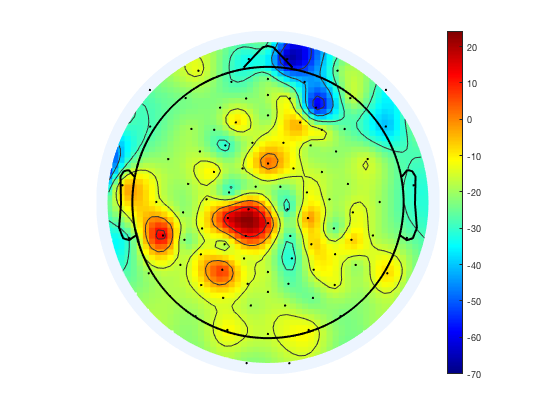}
\caption{NS$_i$}
  \label{fig:11}
\end{subfigure}
\begin{subfigure}{0.15\textwidth}
\centering
\includegraphics[width=\linewidth, trim={2cm 1.5cm 2cm 1cm},clip]{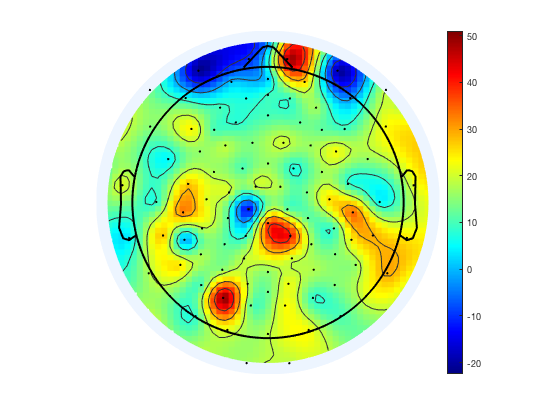}
\caption{NS$_e$}
  \label{fig:11}
\end{subfigure}
\caption{Average Topographic map across trials of one session}
  \label{fig:topoerp}
  \vspace{-0.4cm}
\end{figure}

To cross-validate our findings, we evaluate our methods by visual discrimination to detect distinct signatures of NS classes, if any. The temporal STE segments corresponding to the 3 sub-divisions of NS are plotted using the CED approach. A t-SNE plot is also generated by considering 20-sample chunks of data from the class segments as one data instance. While Figure \ref{fig:vis} shows good discrimination between the beginning and ending NS states, some of the NS$_e$ get classified as NS$_i$(possibly due to the subject's uncertainty about the phrase' end point). The topographic map analysis in Figure \ref{fig:topoerp}, further shows similarity in the NS$_i$ and NS$_e$ states as opposed to the NS$_b$ state thus strengthening the claims mentioned above. The strong modelling of the NS$_b$ state could be due to the brain activation due to anticipation as studied in \cite{leaver2009brain}.

\section{Conclusion}\label{sec:conc}
This work proposes a hierarchical classification approach using an activity detection stage which distinguishes brain activity regions during heard and imagined speech, followed by a search-space reduced unit classification module. Using 4-channel Muse and 128-channel EGI, it is established that the proposed architecture significantly outperforms the baseline methods. Temporal and spatial visual representations further support the reported observations. In conclusion, this work explores the brain responses to speech silences using non-invasive recordings and applies the learning to achieve the broad objective of improved imagined EEG unit classification.  


\clearpage
\bibliographystyle{IEEEtran}
\bibliography{mybib}

\end{document}